\documentclass{article} 
\usepackage{iclr2017_workshop,times}
\usepackage{hyperref}
\usepackage{url}
\pdfoutput=1
\usepackage{natbib}
\usepackage{amssymb}
\usepackage{amsmath}
\usepackage{textcomp}
\usepackage{siunitx}
\usepackage{graphicx}
\usepackage{caption}
\usepackage{subcaption}
\usepackage{booktabs}
\usepackage{multirow}
\usepackage{tabularx}
\title{Learning to Predict Where to Look in Interactive Environments Using Deep Recurrent Q-Learning}

\author{Sajad Mousavi, Michael Schukat, Enda Howley\\
The College of Engineering and Informatics\\
National University of Ireland, Galway\\
\texttt{\{s.mousavi1,michael.schukat,ehowley\}@nuigalway.ie}\\
\And
Ali Borji\\
Department of Computer Science\\
University of Central Florida (UCF)\\
Orlando, Florida 32816-2365, USA \\
\texttt{aborji@crcv.ucf.edu}\\
\And
Nasser Mozayani \\
School of Computer Engineering\\
Iran university of Science and Technology (IUST)\\ 
Tehran, Iran \\
\texttt{mozayani@iust.ac.ir} \\
}

\begin{document}

\maketitle

\begin{abstract}
Bottom-Up (BU) saliency models do not perform well in complex interactive environments where humans are actively engaged in tasks (e.g., sandwich making and playing the video games). In this paper, we leverage Reinforcement Learning (RL) to highlight task-relevant locations of input frames. We propose a soft attention mechanism combined with the Deep Q-Network (DQN) model to teach an RL agent how to play a game and where to look by focusing on the most pertinent parts of its visual input. Our evaluations on several Atari 2600 games show that the soft attention based model could predict fixation locations significantly better than bottom-up models such as Itti-Koch’s saliency and Graph-Based Visual Saliency (GBVS) models.
\end{abstract}

\section{Introduction}

Human visual attention is attracted either by salient stimuli reflected from the environment or task demands where an observer is attracted to objects of interest \citep{Connor+Egeth2004}. Researchers have noticed that human visual system does not perceive and process the whole visual information provided at once. Instead, humans selectively pay attention to different parts of the visual input to gather relevant information sequentially and try to combine information from each step over time to build an abstract representation of the entire input \citep{Connor+Egeth2004,Rensink2000}. Works done in neuroscience and cognitive science literature have revealed selection of regions of interest is highly dependent on two modes of visual attention: bottom-up and top-down. The bottom-up mode attends to low-level features of potential importance and attention is represented in the form of a saliency map \citep{Itti+Koch1998,Harel+Koch2006}. The top-down mode deals with task demands and goals which strongly influence scene parts to which humans should fixate \citep{Torralba2006,Hayhoe2005}. Attention models inspired by the human perception have shown good results in a variety of applications of computer vision such as object recognition and detection \citep{Navalpakkam2005,Walther2006}, video compression  \citep{Itti2004,Guo2010} and virtual reality \citep{Seif2009}. Moreover, with the increased interest in deep learning paradigm in recent years, visual attention mechanisms incorporated in these methods have also shown astonishing results in a wide range of applications including, image captioning \citep{Lin2014,Xu2015}, machine translation \citep{Bahdanau2014}, speech recognition and object recognition \citep{Gonzalez2015}.

Reinforcement learning \citep{Sutton98} is one of the most powerful frameworks in solving sequential decision making problems, where decision making is based on a sequence of observations of the task’s environment. Until recently, applying reinforcement learning to some real world applications where the input data is high dimensional (e.g., vision and speech) was a major challenge. Recent advances in deep learning has resulted in powerful tools for automatic feature extraction from raw data, e.g. raw pixels of an image. Research has shown that deep neural networks can be combined with reinforcement learning in order to learn useful representations. For example, Deep Q-Network (DQN) algorithm \citep{Mnih2013,Mnih2015}, which is a combination of Q-learning with a deep neural network, has achieved good performances on several games in the Atari 2600 domain and in some games it can gain even higher scores than the human player. This combination, deep neural networks and reinforcement learning framework, has also been shown to achieve promising results in the computer vision domain \citep{Ba2014,Sermanet2014}, specifically in visual attention based models. The main challenge in sequential attention models is leaning where to look. Reinforcement learning techniques such as policy gradients are good choices to address this challenge \citep{Mnih2014}.

\begin{figure}[!tbp]
  \begin{subfigure}[b]{0.65\textwidth}
    \includegraphics[width=\textwidth, height=6cm]{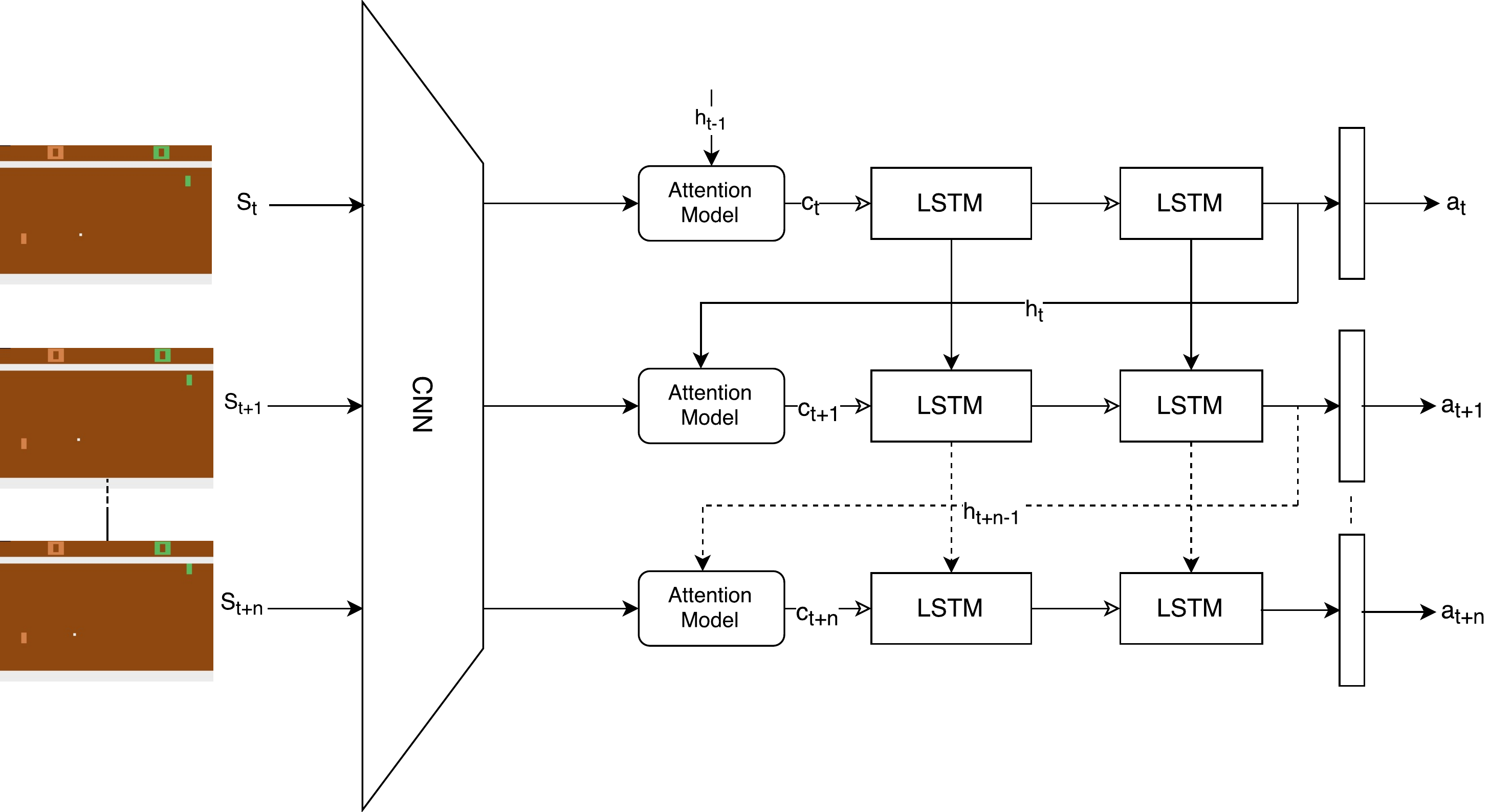}
    \caption{}
    \label{fig:f1}
  \end{subfigure}
  \hfill
  \begin{subfigure}[b]{0.3\textwidth}
    \includegraphics[width=\textwidth]{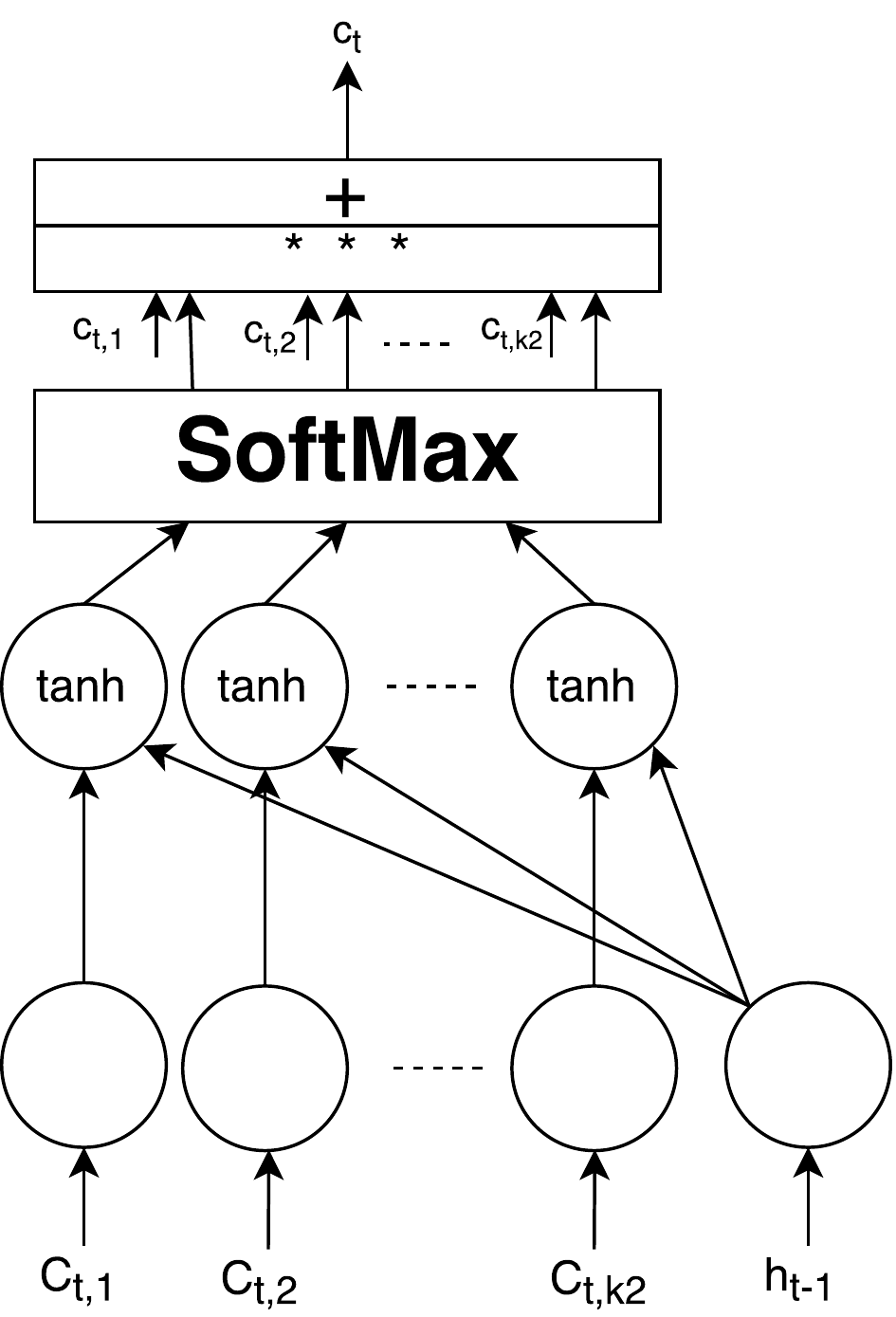}
    \caption{}
    \label{fig:f2}
  \end{subfigure}
  \caption{ (\ref{fig:f1}) Architecture of our model (a deep Q-network with possibility of visual attention). The CNN part of the model takes as input a sequence of states of the game and extracts feature maps. It then computes vertical feature slices, $C_t$ with dimension $D$. (\ref{fig:f2}) The attention model. at each time step $t$, the attention model uses $C_t$ along with the previous internal state of the LSTM part, $h_{t-1}$ to produce an expected value, $c_t$ of vertical feature slices, $C_{t,i}$ regarding $\alpha_{t,i}$, the importance of each part of input frame.}
  \label{fig:model}
\end{figure}

Overall, there are two types of attention models, the soft attention models and the hard attention models. The soft attention models are end-to-end approaches and differentiable deterministic mechanisms that can be learned by gradient based methods. However, the hard attention models are stochastic processes and not differentiable. Thus, they can be trained by using the REINFORCE algorithm \citep{Williams1992} in the reinforcement learning framework. In this paper, a soft attention mechanism is integrated into the deep Q-network (as shown in Figure \ref{fig:model}) to enable a deep Q-learning agent to learn to play Atari 2600 games by focusing on the most pertinent parts of its visual input. In fact, the model tries to learn the control actions and the attention locations simultaneously.
 
To test our model, we compare predicted fixation locations with explicit attention judgements of people with a specific scenario as explained in section \ref{psy_coll}. The results give better fixation prediction accuracy compared to two popular bottom-up (BU) saliency models: Itti-Koch’s saliency model \citep{Itti+Koch1998} and Graph-Based Visual Saliency (GBVS) model \citep{Harel+Koch2006}. We also demonstrate that the proposed model can learn how to play Atari 2600 games (i.e. leaning the control actions of the game) and where to look (i.e. learning the fixation locations) on video game frames effectively.

\section{Related Work}

Experiments studying eye movements in natural behaviour have proved that the task plays a strong role in learning where and when human fixate, and that the eyes are guided to the points that are sometimes non-salient \citep{Torralba2006,Hayhoe2005,Gonzalez2015}. For instance, some works on daily activities such as driving \citep{Land1994}, tea making \citep{Land2001} and making a sandwich \citep{Hayhoe2005} have demonstrated that almost all fixations fall on task-relevant objects and very few irrelevant regions are fixated. Recognizing the influence of reward signals \citep{Schultz2000} on eye movements and in directing eye gaze, some researchers (e.g., \citet{Sprague2003}) have shown that reinforcement learning is useful for modelling eye movement. They suggested an RL-based method to learn visio-motor behaviours by considering uncertainty costs when an eye movement action is made in a sidewalk navigation task of a virtual urban environment \citep{Sprague2007}, where the goal is maximizing discounted sum of future rewards (or at least cost) over time.

Combining deep learning and reinforcement learning has achieved impressive results in learning to play various video games \citep{Mnih2013,Mnih2015,VanHasselt2015,Guo2014,Silver2016} as well as in different problems of computer vision field. For instance, combining deep Convolutional Neural Networks (CNNs) with multi-layered Recurrent Neural Networks (RNNs) in particular, with Long Short-Term Memory (LSTM) components that use reinforcement learning, have led to end-to-end systems for deciding where to look. Such techniques could perform well in classification and recognition tasks, e.g. recurrent  neural models of attention for fine-grained categorization on the Stanford Dogs data set \citep{Sermanet2014}, cluttered digit classification on the  MNIST date set and playing a toy visual control problem \citep{Mnih2014}, and the recognition of house number sequences from Google Street View images \citep{Ba2014}.

Both soft attention and hard attention mechanisms have gained remarkable performance over different challenging problems. Recent research done by \citet{Xu2015} has applied both mechanisms to generate image captions. The attention mechanism in this work aims to generate each word of the caption by focusing on the relevant parts of the image. Another approach using an attention mechanism is called spatial transformer networks \citep{Jaderberg2015}. Spatial transformer networks have used affine transformations as a soft attention mechanism to focusing on the relevant part of an image. More recently,  \citet{Sharma2015} used a recurrent soft attention based model for action recognition in videos and showed that their model performs better than some baselines (without the attention mechanism).  In this paper, inspired by the work of \citet{Sharma2015}, we propose a soft attention mechanism incorporated in a Deep Q-Network \citep{Mnih2013,Mnih2015} to learn where to look and which actions to select. \citet{Wang2015} have proposed a new neural network architecture to play Atari 2600 games, which is an extension of the Double Deep Q-Network (Double DQN) in \citep{VanHasselt2015}. Applying their method to several Atari 2600 games, they obtained state-of-the-art results. They also visualized saliency maps, indicating where the agent looks at, while taking an action. To this end, they generated the saliencies by computing the Jacobians of the state-value and action-value functions, inspired by the way introduced in \citep{Simonyan2013}. In contrast, here, we use multi-layered LSTMs and the soft attention mechanism to predict fixation locations and compare our model’s accuracy with the bottom-up saliency models. 
\section{Model Description}
\label{mod_des}
\subsection{Model Architecture: ConvNet + Attention Model + LSTM}
\label{mod_des}
The idea behind the proposed model is similar to the work done in \citep{Sharma2015}, but with differences in terms of the way the model is trained (i.e. reinforcement learning) and the application domain. Here we are interested in learning to play Atari 2600 games by selecting the appropriate control actions provided for each game while paying attention to the most important locations. 

Our model is an aggregation of CNN, soft attention mechanism and RNN (i.e. LSTM layers), as shown in Figure \ref{fig:model}. In order to extract convolutional features, we used the first three layers of the Deep Q-Network (DQN) \citep{Mnih2015} as the CNN part of the model, three stacks of convolutions plus rectifier nonlinearity layers with filters $32\ 8\times8$, $64$ $4\times4$ and $64\ 3\times3$, and strides $4, 2$ and $1$, respectively. At each time-step $t$, a visual frame is fed into the system and the last convolutional layer of the CNN part outputs $D$ feature maps $K \times K$ (for example, here $64$ feature maps $7\times7$). Then, feature maps are converted to $K^2$ vectors in which each vector has $D$ dimension as follows: $$C_t = [C_{t,1},C_{t,2}, \ldots ,C_{t,K^2}], \qquad C_{t,i}\in \mathbb{R}^D.$$
Next, they are fed to the attention model to compute the probabilities corresponding to the importance of each part of the input frame. In other words, the input frame is divided into $K^2$ regions and the attention mechanism tries to attend to the most relevant region.

We utilize the soft attention mechanism of \citet{Xu2015} (See \citet{Bahdanau2014,Cho2015} for a detailed discussion). Figure \ref{fig:f2} shows the structure of the attention mechanism. The attention model takes $K^2$ vectors, $C_{t,1},C_{t,2}, \ldots ,C_{t,K^2}$ and a hidden state $h_{t-1}$. It then produces a vector $c_t$ which is a linear weighted combination of the values of $C_{t,i}$. The $h_{t-1}$ is the internal state of the LSTM at the previous time step. Each vector $C_{t,i}$ is a representation of different regions of the input frame. More formally, the attention module tries to select the vectors using a linear combination of  $h_{t-1}$ and $C_{t,i}$, i.e:
\begin{align}
&c_t=E[C_t]=\sum_{i=1}^{k \times k} \alpha_{t,i} C_{t,i}, \\
&\alpha_{t,i}= \frac{\exp({f_{att}(C_{t,i},h_{t-1}}))}{\sum_{j=1}^{k \times k} \exp({f_{att}(C_{t,i},h_{t-1}}))} \qquad i \in 1,2,\ldots ,{k \times k}, \\
&f_{att}(C_{t,i},h_{t-1})=\tanh({W_{hatt}}^T h_{t-1}+{W_{catt}}^T C_{t,i}),
\end{align}
At each time step $t$, attention module computes $f_{att}$, a $\tanh$ layer which is a composition of the values of $C_{t,i}$ and $h_{t-1}$. Then uses it to calculate $\alpha_{t,i}$, a softmax over ${k \times k}$ regions. They can be considered as the amount of the importance of the corresponding vector $C_{t,i}$ among of ${k \times k}$ vectors in the input image. Once $\alpha_{t,i}$ values are ready, the attention model calculates $c_t$, a weighted sum of all vectors $C_{t,i}$ based on given $\alpha_{t,i}$s. Thus, in this way, the RL agent can learn to emphasize on the interesting part of the input frame based on the given state. Note that the soft attention model is fully differentiable  which allows training the systems in an end-to-end manner.

The RNN part of the network, which is a stack of two LSTM layers with the LSTM sizes 64, uses the previous hidden state $h_{t-1}$ and the output of the attention module $c_t$ to calculate the next hidden state $h_{t}$. The $h_{t}$ is used as input of the output layer, which is a fully-connected linear layer with the output neurons for each legal action of the game. It  is also utilized as input of the attention module in order to calculate the value $c_{t+1}$ at the next time-step.

\subsection{Model Training}
\label{mod_training}
The goal of the RL agent is to learn an optimal policy $\pi$,  the probability of selecting action $a$ in state $s$ by focusing on the most relevant part of the given state, such that by following the underlying policy the sum of the discounted rewards over time is maximized. Since the proposed model is an end-to-end learning system, all of the network parameters (i.e. three parts of the model: ConvNet, the attention network and the LSTMs), $\theta_t$ can be learned by trying to minimize the following loss function of mean-squared error in Q values:
\begin{align}
&L(\theta)=E[(r+\gamma max_{a^\prime}Q(s^\prime,a^\prime;\theta_{t-1})-Q(s,a;\theta_t))^2],
\end{align}
where $r+\gamma max_{a^\prime}Q(s^\prime,a^\prime;\theta_{t-1})$ is the target value, $r$ is a scalar reward which the agent receives after taking action $a$ in state $s$. Parameter $0 \leqslant \gamma \leqslant 1$ is called the discount factor. For optimizing the above loss function, we utilize the stochastic gradient descent method. Thus, in the Q-learning algorithm, the parameters are updated as follows:
\begin{align}
&\theta_i= \theta_{i-1}+ \alpha{(y_i-Q(s,a;\theta_i))} \frac{\partial Q(s,a;\theta_i)} {\partial \theta_i},
\end{align}
where it is implicit that $y_i=r+\gamma max_{a^\prime}Q(s^\prime,a^\prime;\theta_{t-1}$) is the target value for iteration $i$, and $\alpha$ is a learning rate.
\begin{figure}[!tbp]
  \begin{subfigure}[b]{\textwidth}
  \textbf{\qquad Breakout \qquad\qquad Enduro \qquad\qquad Phoenix\qquad\qquad \qquad Pong\qquad\qquad Seaquest   }\par
    \includegraphics[width=\textwidth]{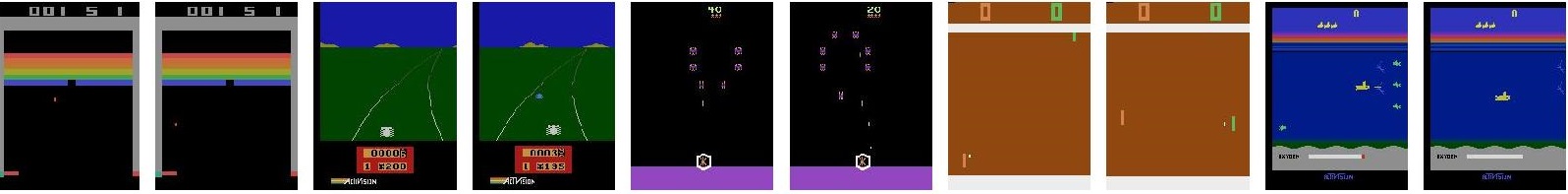}
     \end{subfigure}
  \hfill
  \begin{subfigure}[b]{\textwidth}
    \includegraphics[width=\textwidth]{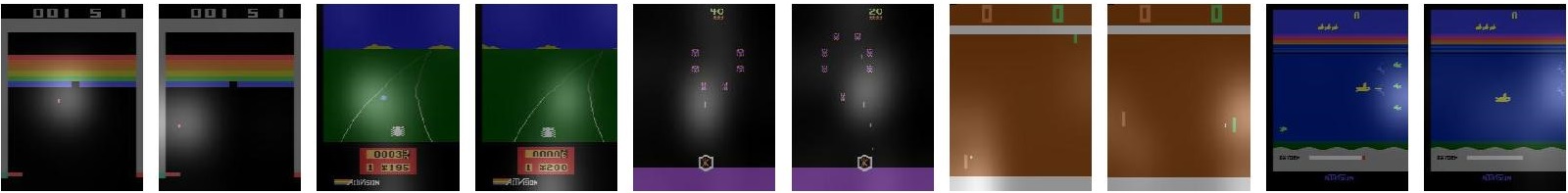}
  \end{subfigure}
  \caption{ Visualization of attended locations by the proposed model over sample frames of the five Atari 2600 games. The white regions of frames indicate where the model is interested to look at. Higher brightness means more attention.}
  \label{fig:samp_fr}
\end{figure}
\section{Experiment and Results}
\label{exp_res}
\subsection{Evaluation metrics and Hyperparameters}
To compare the power of the proposed model for predicting fixation locations against the other models, we used two metrics: Normalized Scanpath Saliency (NSS) \citep{Peters2005} and Area Under Curve (AUC) \citep{Bruce2005}.

NSS is used for evaluating the salience map values using fixated locations. It is a response value at the human eye position $(x_h,y_h)$ of the saliency map ($sm$) suggested by a model, where saliency scores of the predicted map density have zero mean and unit standard deviation. NSS can be formulated as follows:
\begin{align}
&NSS= \frac{sm(x_h,y_h)-\mu_{sm}}{\sigma_{sm}},
\end{align}
where  ${\sigma^2}_{sm}$ and $\mu_{sm}$ are the variance and the mean of the saliency map ($sm$).

NSS equal 1 means that the subject’s eye position fall in a region where the predicted map density is one standard deviation above average, while NSS $=$ 0 suggests the model would not perform better than random. NSS $<$ 0 indicates attention to non-salient locations predicted by the model.

AUC is the area under the Receiver Operating Characteristics (ROC) curve \citep{Bruce2005}.  It is one the most commonly used ways to evaluate the performance of a binary classifier. The ROC curve is created based on the true positive rate (TPR) against the false positive rate (FPR) for different cut-off points of a parameter (here, the salience value). Each pair (TPR, FPR) of the ROC curve is computed with a particular threshold. To use this metric in this work, human fixations on the image are considered as ground truth and the saliency map can be viewed as a binary classifier to classify each pixel of the image to fixated and non-fixated samples. By drawing the ROC curve and calculating the area under the ROC curve, we can evaluate how well the saliency map predicts actual human eye fixations. The perfect predication has a score of 1.0, while a score of 0.5 indicates that the model is no better chance.

In all of our experiments, we trained the system with a 2-layer LSTM with 64 hidden/cell units and a value of 4 for the number of sequence steps in backpropagation through time. In order to ensure stable learning, LSTM gradients were clipped to a value of 10 \citep{Hausknecht2015}. At each step of training the network, the initial LSTM hidden and cell states were set to zero. Exploration strategy of the agent was $\epsilon$-greedy policy with $\epsilon$ decreasing linearly from 1 to 0.1 over the first million steps. The discount factor was set to $\gamma = 0.99$. The networks were trained for 2 million steps and the size of the replay memory was 500, 000. All network weights were updated by the RMSProp optimizer \citep{Yu2010} with mini batches of size 32, a momentum of 0.95, and a learning rate of $\alpha = 0.00025$. Training for all the games was done with the same network architecture and hyperparameters.
\begin{figure}[!tbp]
  \begin{subfigure}[b]{\textwidth}
  \textbf{\qquad Original frame \qquad\qquad Itti-Koch \qquad\qquad\qquad \quad GBVS\qquad\qquad\qquad RL-based}\par
    \includegraphics[width=\textwidth, height=3.5cm]{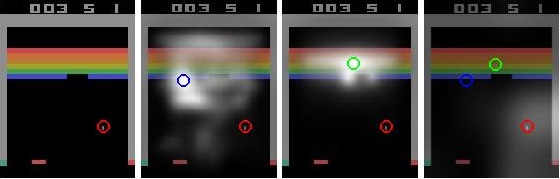}
     \end{subfigure}
  \hfill
  \begin{subfigure}[b]{\textwidth}
    \includegraphics[width=\textwidth, height=3.5cm]{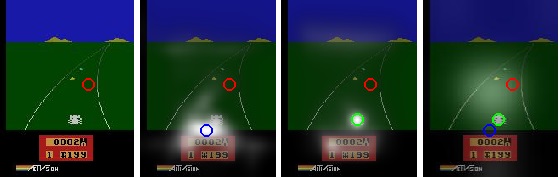}
  \end{subfigure}
    \hfill
  \begin{subfigure}[b]{\textwidth}
    \includegraphics[width=\textwidth, height=3.5cm]{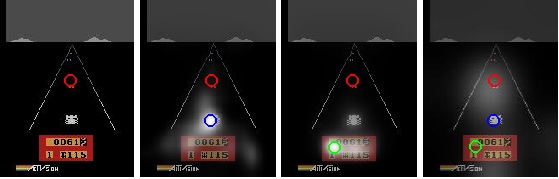}
  \end{subfigure}
    \hfill
  \begin{subfigure}[b]{\textwidth}
    \includegraphics[width=\textwidth, height=3.5cm]{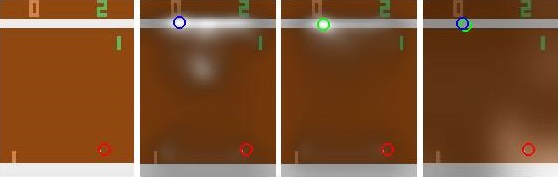}
  \end{subfigure}
    \hfill
  \begin{subfigure}[b]{\textwidth}
    \includegraphics[width=\textwidth, height=3.5cm]{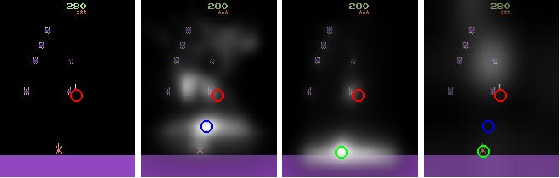}
  \end{subfigure}
    \hfill
  \begin{subfigure}[b]{\textwidth}
    \includegraphics[width=\textwidth, height=3.5cm]{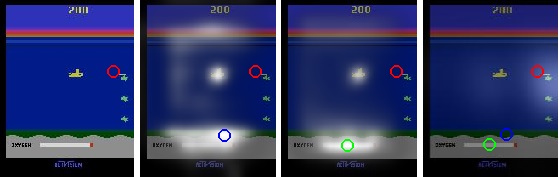}
  \end{subfigure}
  \caption{Each row shows a sample frame of an Atari 2600 game and corresponding saliency maps overlaid on the frame from different models. The red circles indicate the human fixation (i.e., the clicked positions by the subjects). The green and blue circles indicate the location of the maximum point in each map for GBVS and Itti-Koch algorithms, respectively. The white parts indicate attended regions by the model. Higher brightness means more attention.}
  \label{fig:samp_sail}
\end{figure}
\begin{figure}
\centering
  \includegraphics[width=0.8\linewidth]{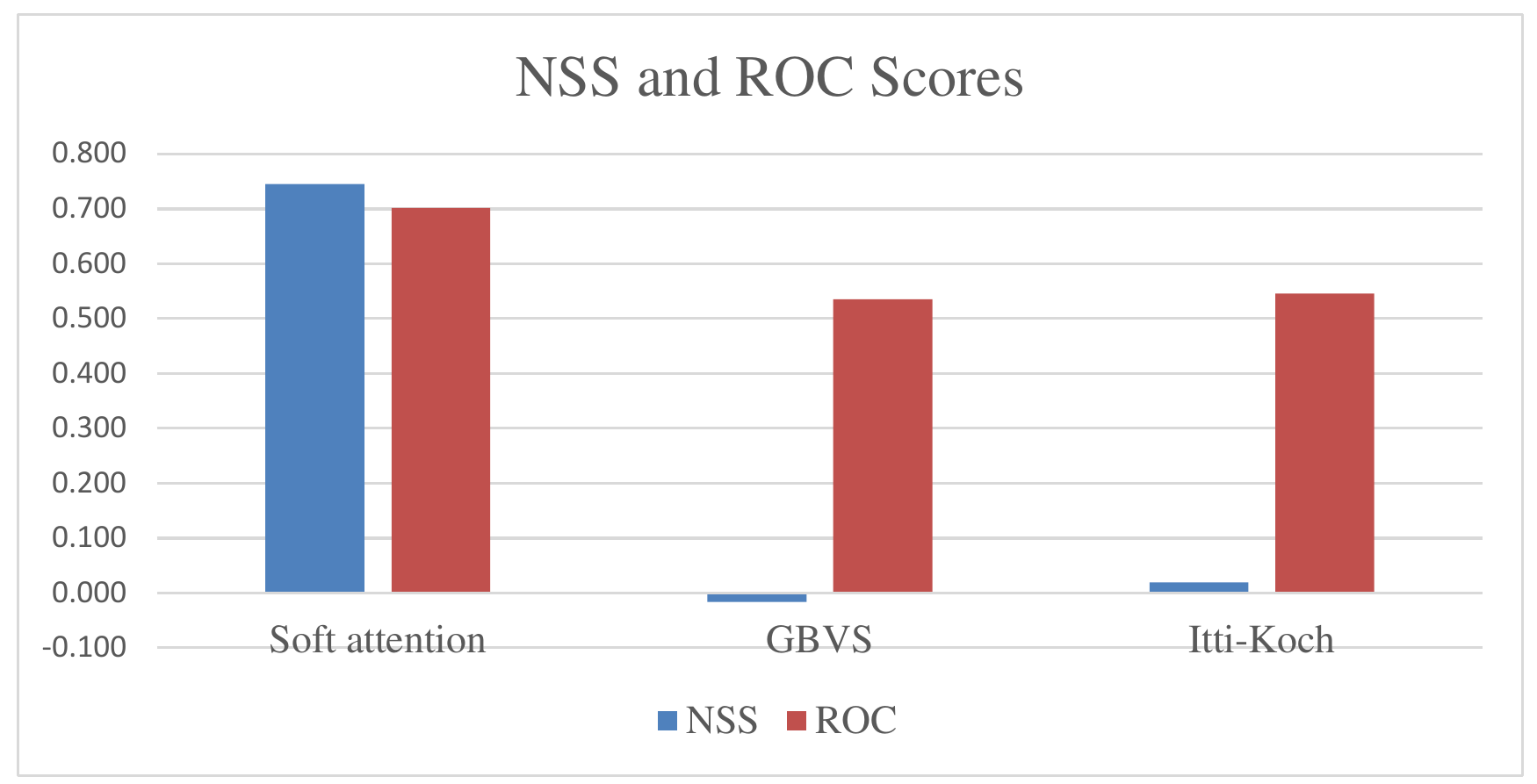}
  \caption{NSS and ROC scores of predicted fixation locations via the different models.}
  \label{fig:chart}
\end{figure}

\subsection{Experiment Setup}
\subsubsection{The Arcade Learning Environment}
We have used Atari 2600 games in the Arcade Learning Environment (ALE) \citep{Bellemare2013}. The ALE provides an environment that emulates the Atari 2600 games. It presents a  very  challenging environment  for  the reinforcement  learning and other approaches that  have a  high dimensional  visual  input ($210 \times 160$ RGB video at 60Hz),  which might also be partially observable. It presents a wide range of interesting games that are useful to be a standard testbed for evaluating the novel algorithms. In our experiments, we used 5 Atari games: Pong, Phoenix, Enduro, Breakout and Seaquest. Our goal is training a RL agent to learn a certain optimal policy to play each of the games and attention to task-relevant parts of the input frame. In fact, learning to choose the optimal action based on the region selected from among provided candidates (vertical feature slices resulting from the last convolution layer of the CNN part of the model, see $C_t$ in section \ref{mod_des}). Figure \ref{fig:samp_fr} shows some sample frames of the utilized games as well as the learned attended regions.

\subsubsection{Psychophysics and Collecting Data}
\label{psy_coll}
To test the accuracy of the model, we recruited three subjects (two males and one female). They were not experts in playing Atari 2600 games. So, at first we provided them five games (Pong, Phoenix, Enduro, Breakout and Seaquest) of the set of Atari 2600 games. We asked subjects to play the games to learn how to play and familiarize themselves with the environment. Second, we showed the recorded videos played by the RL agent to subjects (the frame rate: 5fps; higher rate was a little fast for the subjects to concentrate). Subjects were asked to click at frame locations where they thought should be looked at while actually playing the games. Then, we collected the clicked positions and corresponding frames as the ground truths (surrogates for actual eye positions of the subjects) in our analyses. For each game, we considered one episode, which is a trajectory from initial to a terminal state.

\subsubsection{Bottom-up saliency models}
In order to evaluate the suggested model, we compared the performance of our model against two popular bottom-up saliency (BU) models: Itti-Koch’s saliency model \citep{Itti+Koch1998} and Graph-Based Visual Saliency (GBVS) introduced by \citet{Harel+Koch2006}. We used the freely available implementations of the saliency models provided by Harel\footnote{J. Harel, A saliency implementation in MATLAB: \url{http://www.klab.caltech.edu/~harel/share/gbvs.php}}. Itti-Koch’s bottom up model uses twelve features extracted from the image, including color contrast (red/green and blue/yellow), temporal luminance flicker, luminance contrast, four orientations (\ang{0}, \ang{45}, \ang{90}, \ang{135}), and four oriented motion energies (up, down, left, right). Next, it computes conspicuity maps of individual features through center surround architecture inspired by biological receptive fields. It then linearly combines saliency from different features to generate a unique master saliency map representing the conspicuity of each location in the visual input. The GBVS algorithm attempts to create feature maps in the way done by Itti-Koch’s method, but normalizes them using a graph based approach to reach a better combination of conspicuous maps.
\begin{table}[htbp]
  \centering
  \caption{Comparison of performance of the proposed model against bottom-up models on several frames of the five Atari 2600 games}
\begin{tabular}{llllll|llll}
\cline{2-7}
\multicolumn{1}{l|}{} & \multicolumn{2}{c|}{\textbf{Breakout (724 frames)}} & \multicolumn{2}{c|}{\textbf{Enduro (2914 frames)}} & \multicolumn{2}{c|}{\textbf{Phoenix (575 frames)}} &  &  &  \\ 
\cline{1-7}
\multicolumn{1}{|l|}{\textbf{Model}} & \multicolumn{1}{l|}{NSS} & \multicolumn{1}{l|}{ROC} & \multicolumn{1}{l|}{NSS} & \multicolumn{1}{l|}{ROC} & NSS & \multicolumn{1}{l|}{ROC} &  &  &  \\ 
\cline{1-7}
\multicolumn{1}{|l|}{Soft attention} & \multicolumn{1}{l|}{1.326} & \multicolumn{1}{l|}{0.787} & \multicolumn{1}{l|}{0.699} & \multicolumn{1}{l|}{0.689} & 0.187 & \multicolumn{1}{l|}{0.529} &  &  &  \\ 
\cline{1-7}
\multicolumn{1}{|l|}{GBVS} & \multicolumn{1}{l|}{-0.074} & \multicolumn{1}{l|}{0.489} & \multicolumn{1}{l|}{-0.083} & \multicolumn{1}{l|}{0.578} & -0.143 & \multicolumn{1}{l|}{0.485} &  &  &  \\ 
\cline{1-7}
\multicolumn{1}{|l|}{Itti-Koch} & \multicolumn{1}{l|}{-0.112} & \multicolumn{1}{l|}{0.453} & \multicolumn{1}{l|}{-0.048} & \multicolumn{1}{l|}{0.593} & 0.267 & \multicolumn{1}{l|}{0.599} &  &  &  \\ 
\cline{1-7}
 &  &  &  &  & \multicolumn{1}{l}{} &  &  &  &  \\ 
\cline{3-6}
 & \multicolumn{1}{l|}{} & \multicolumn{2}{c|}{\textbf{Pong (1346 frames)}} & \multicolumn{2}{c|}{\textbf{Seaquest (577 frames)}} &  &  &  &  \\ 
\cline{2-6}
\multicolumn{1}{l|}{} & \multicolumn{1}{l|}{\textbf{Model}} & \multicolumn{1}{l|}{NSS} & \multicolumn{1}{l|}{ROC} & \multicolumn{1}{l|}{NSS} & ROC &  &  &  &  \\ 
\cline{2-6}
\multicolumn{1}{l|}{} & \multicolumn{1}{l|}{Soft attention} & \multicolumn{1}{l|}{0.846} & \multicolumn{1}{l|}{0.76} & \multicolumn{1}{l|}{0.571} & 0.694 &  &  &  &  \\ 
\cline{2-6}
\multicolumn{1}{l|}{} & \multicolumn{1}{l|}{GBVS} & \multicolumn{1}{l|}{0.179} & \multicolumn{1}{l|}{0.478} & \multicolumn{1}{l|}{0.058} & 0.564 &  &  &  &  \\ 
\cline{2-6}
\multicolumn{1}{l|}{} & \multicolumn{1}{l|}{Itti-Koch} & \multicolumn{1}{l|}{0.129} & \multicolumn{1}{l|}{0.472} & \multicolumn{1}{l|}{0.023} & 0.54 &  &  &  &  \\ 
\cline{2-6}
\end{tabular}
  \label{tab:tab_NSS_ROC}
\end{table}
\subsection{Results}
We ran the two bottom-up saliency models (i.e. Itti-Koch saliency model and the GBVS algorithm) over five Atari 2600 games and compared their NSS and AUC scores to scores achieved by our suggested model. Table \ref{tab:tab_NSS_ROC} reports a comparison of the proposed model with those bottom-up models. The results on Table \ref{tab:tab_NSS_ROC} demonstrate that the soft attention model significantly performs better than the Itti-Koch and the GBVS methods using both evaluation metrics over all five games. We observe that NSS and ROC scores for some games like Breakout, Enduro and Pong are better than others. Further, movements of objects (especially, enemies) in the games Phoenix and Seaquest are high compared to the other games which makes learning an appropriate control policy difficult. As a result to better predict fixation locations, the algorithm needs more than 2 million time steps. For fair comparison, similar to the other games, we stopped training the networks of these games after 2 million steps.  

Figure \ref{fig:samp_sail} shows several sample frames of five Atari games and their corresponding saliency maps overlaid on the frames from the investigated models. The distance between interesting places of the frame for the subject (shown in red circles in the figure) and predicted fixation locations (i.e. the location of the maximum point in the saliency map) of the models (shown in blue and green circles) illustrates how well a model’s predictions match subjects’ click positions. We see that the soft attention model (RL-based model) is able to suggest significantly better attention predictions compared to the GBVS and Itti-Koch methods. 

Figure \ref{fig:chart} shows the NSS and ROC scores of the model across all frames of five games. The plot illustrates a big performance difference between bottom-up saliency models, and the proposed model. For instance, the soft attention model obtains mean NSS and ROC scores of 0.74 and 0.70 while bottom-up saliency models do not perform better than random. From these results, we learn that bottom-up saliency models performs poorly in interactive tasks where humans are actively engaged in tasks (e.g. playing video games). The interested readers are referred to \citet{Borji2014}, where they propose several top-down attention models and exhaustively compare them with bottom-up saliency models.

\section{Conclusion}
\label{con}
In this paper, we utilized a combination of Deep Q-network algorithm and recurrent soft attention mechanism to decide where to look as well as learn action selection in complex interactive environments. We applied the proposed model to predict fixation locations, while a reinforcement learning agent is playing Atari 2600 games. To show fixation prediction accuracy of our model, we set up a scenario to collect click locations (as measure of explicit attention) from a number of subjects while they are watching the videos of played games. We ran the two most commonly used bottom-up saliency models over five recorded video game video clips which have been played by a RL agent. We compared the model to those bottom up models. Our experiments show that saliency maps suggested by our model obtain significantly better AUC and NSS scores. Despite the remarkable results, we are already feeding the entire frame into the system, which is computationally expensive. Using only focused parts of the given frame as input of the system (i.e., foveated representation) would be an interesting future research direction.

\bibliography{iclr2017_workshop}
\bibliographystyle{iclr2017_workshop}

\end{document}